# aHUGIN: A System Creating Adaptive Causal Probabilistic Networks

**Kristian G. Olesen** **Steffen L. Lauritzen** **Finn V. Jensen**
Institute for Electronic Systems
Aalborg University
Fredrik Bajers Vej 7
DK-9220 Aalborg
Denmark

## Abstract

The paper describes aHUGIN, a tool for creating adaptive systems. aHUGIN is an extension of the HUGIN shell, and is based on the methods reported by Spiegelhalter and Lauritzen (1990*a*). The adaptive systems resulting from aHUGIN are able to adjust the conditional probabilities in the model. A short analysis of the adaptation task is given and the features of aHUGIN are described. Finally a session with experiments is reported and the results are discussed.

## 1 Introduction

With the revival of Bayesian methods in decision support systems (Shachter 1986; Pearl 1988; Shafer and Pearl 1990; Andreassen *et al.* 1991*b*) mainly due to the construction of efficient methods for belief revision in causal probabilistic networks (Pearl 1988; Lauritzen and Spiegelhalter 1988; Andersen *et al.* 1989; Jensen *et al.* 1990; Shenoy and Shafer 1990), the process of knowledge acquisition under the Bayesian paradigm has become increasingly important. When constructing causal probabilistic network models, various sources may be used, ranging from ignorance over experts' subjective assessments to well established scientific theories and statistical models based on large databases. Very often a model is a mixture of contributions from sources of different epistemological character.

Sometimes these contributions do not coincide, and the model is a mediation between them; sometimes the resulting model is incomplete (ignorance has, for example, forced crude 'guesses' on certain distributions); sometimes the model must vary with contexts which cannot be specified beforehand; sometimes the domain is drifting over time — requiring the model to drift along with it, and sometimes the model quite simply does not reflect the real world properly.

All the problems listed above call for procedures which enable the system to modify the model through experience. We call such an activity *adaptation*, and systems performing automatic adaptation we call *adaptive* systems.

Note that we have chosen to distinguish adaptation from *training*, which we use to describe the activity of creating models by batch-processing of large data bases. In Spiegelhalter *et al.* (1992) both activities are called learning and they are distinguished as sequential learning and batch learning. When using adaptation we are using the analogy to the notion of adaptive regulators in control theory. Hopefully, this abundance of terminology will not confuse the reader completely.

The present paper describes aHUGIN, a tool for creating adaptive systems. The system, which is an extension of HUGIN (Andersen *et al.* 1989), is based on methods reported in Spiegelhalter and Lauritzen (1990*a*), see also Spiegelhalter and Lauritzen (1990*b*), and the adaptive systems resulting from aHUGIN are able to adjust the conditional probabilities in the model. In aHUGIN the model is compactly represented by a contingency table of imaginary counts, and the adaptation procedure is a process of modifying the counts in this table.

In section 2 we give a short analysis of the adaptation task and discuss various simple adaptation methods leading up to a description of the one used in aHUGIN. In section 3 we describe the features of aHUGIN, and in section 4 a session with experiments is reported and the results are discussed.

Spiegelhalter and Cowell (1992), and Cowell (1992) describe a similar system and results of slightly different experiments. The main difference between this system and their system, is that we allow an extra facility, called *fading*, that makes the system forget the past at an exponential rate, thereby making them more prone to adapt in changing environments.

## 2 Analysis of adaptation

CPN models have both a quantitative and a qualitative aspect. Through the directed arcs, the network reflects the only ways in which variables may have im-pact on each other. The strength of the impact is modelled through conditional probability tables. We shall here describe how the probability tables are modified in the adaptation process.

So, consider a causal probabilistic network into which information on the state of the variables can be entered. If the state of only some of the variables is known, then the probability distributions for the remaining variables are calculated. Each time this is done, we have described a *case*. Now, a large number of cases is at hand, and we want to improve the model by adjusting the conditional probability tables to the set of cases.

## 2.1 Direct modelling of table uncertainty

In Figure 1 (a), the state of the variable $A$ is influenced directly by the states of $B$ and $C$, and the strength of this influence is modelled by $P(A|B,C)$. If the strength is subject to doubt — this may, for example, be due to different estimates from experts, or it may be due to a context influence not modelled (like soil quality of corn fields or genetic disposition for a disease) — then this doubt may be modelled directly by introducing an extra parent, $T$, for $A$ (Figure 1 (b)). This variable can be considered as a type variable modelling, for example, types of context or different experts' assessments. To reflect credibility of the experts or frequencies of the context types, a prior distribution for $T$ could be given. When a case is entered to the CPN, the calculation of updated probabilities will yield a new distribution on $T$, and we may say that the change of these probabilities reflects what we have learnt from the case. This process is called *retrieval* of experience. The new distribution may now be used as prior probabilities for the next case and its impact on the conditional probabilities found by summing out the type variable. This process is called *dissemination* of experience. The technique has, for example, been used in Andreassen *et al.* (1991*a*), where the system contains a model for metabolism in patients suffering from diabetes. Through a type variable, the system adapts to the characteristics of the individual patient.

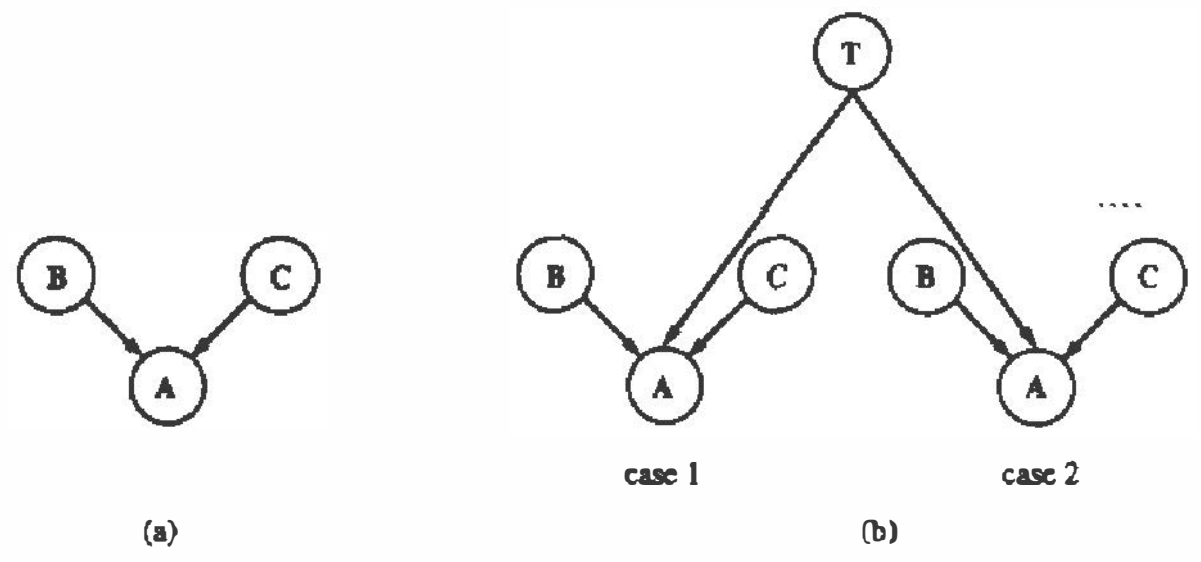

Figure 1: Adaptation through a type variable.

Several conditional probabilities in the CPN may be context dependent, and a whole set of type variables may be necessary. These different types may be heavily interdependent (for example, assessments from various intersecting sets of experts) and it may be necessary to construct a complex type network with the risk of a combinatorial explosion. A simplifying assumption would be *global independence:* the context dependence for the conditional probabilities are mutually independent. In that case, each variable can be given its own parent of types, and retrieval and dissemination are completely local (performed as above). However, the procedure is still vulnerable to combinatorial explosion. Take for instance the variable $A$ in Figure 1. For each parent configuration a type dependence on the distribution for $A$ shall be described. These dependencies may vary a lot with the particular parent configurations and all kinds of inter-dependencies may be present. Therefore we may be forced to increase the probability tables by a factor which is the product of the number of states in the parents. So, a further simplification would be *local independence*: the context dependence for the various parent configurations are mutually independent.

## 2.2 Indirect modelling of table uncertainty

If nothing is known on the structure of the causes for a possible inadequacy of the model to the case set, the uncertainty can not be represented directly through a network of discrete types, and we must leave room for all kinds of types with all kinds of distributions. The learning process here is as everywhere else in the Bayesian paradigm: Specify a prior distribution of the types and calculate the posterior, given the case observed. It remains to find a natural way of specifying such a probability model. Spiegelhalter and Lauritzen (1990*a*) give a range of possibilities, including normal-logistic models.

The simplest probability model which is convenient for this situation assumes that each set of entries in the conditional probability tables for a particular parent configuration follows a so-called *Dirichlet-distribution* (Johnson and Kotz 1972). A $k$-dimensional Dirichlet-distribution has $k$ parameters, $(\alpha_1, \ldots, \alpha_k)$ and density

$$f(p_1, \ldots, p_{k-1}) \propto \left(1 - \sum_{i=1}^{k-1} p_i\right)^{\alpha_k - 1} \prod_{i=1}^{k-1} p_i^{\alpha_i - 1}. \quad (1)$$

for $p_i > 0$ and $\sum_{i=1}^{k-1} p_i < 1$.

The simplicity is in the interpretation: If the (improper) distribution with $\alpha_i = 0$ for all $i$ is considered a noninformative prior, the distribution specified may be interpreted as representing past experience as *a contingency table* $(\alpha_1, \ldots, \alpha_k)$ *of counts of past cases*. The quantity $s = \sum_i \alpha_i$ is therefore referred to as *the equivalent sample size*. The updating procedure consists of modifying the counts as new cases are being observed.

We shall not give details, but just state that the frac-tion $\alpha_i/s = m_i$ is the mean for the $i$th outcome, and for each $i$ the variance of the probability for the $i$th outcome is

$$v_i = \frac{m_i(1-m_i)}{s+1} \tag{2}$$

Hence $v_i$ is a measure of the uncertainty of the probability $m_i$.

Using this interpretation we also have a tool to model expert opinions of the type "the probability is somewhere between $p$ and $q$, but I believe it is about $r$". In the case of two states $a$ and $b$, consider, for example, the statement that the probability of $a$ is between 0.3 and 0.4 and that it is about .35. If we, as in Spiegelhalter *et al.* (1990), interpret the statement so that the mean is 0.35 and the standard deviation is 0.05, then it can be modelled by a 2-dimensional Dirichlet-distribution (which is called a Beta-distribution). We then have to determine two counts $\alpha_a$ and $\alpha_b$ which satisfy the equations

$$\frac{\alpha_a}{\alpha_a+\alpha_b} = 0.35 \text{ and } \frac{0.35\cdot 0.65}{\alpha_a+\alpha_b+1} = 0.0025 \tag{3}$$

which we solve to get $\alpha_a = 31.5$ and $\alpha_b = 58.5$. This can be an attractive alternative to modelling second order uncertainty by intervals of lower and upper probabilities.

Back to learning: Let $(m_1,\ldots,m_n)$ and a sample size $s$ be a given specification of the conditional probability table $P(A|b,c)$. We can then act as if we had a contingency table of counts $(sm_1,\ldots,sm_n)$. If we get a case in the configuration $(b,c)$ and $a_i$, then retrieval quite simply consist in adding 1 to the count for $a_i$, and dissemination is just to calculate the new frequencies. If global and local independence can be assumed the scheme is applicable to all tables.

The scheme only works if both the states of $A$ and its parents are known. In general we may anticipate that the provided evidence, $E$, may leave uncertainty on both the states of $A$ and of its parents.

A naive approach in the general case could be to add a count of $P(a_i,b,c|E)$ to the counts for $a_i$. This scheme is known as *fractional updating* (Titterington 1976). However, the scheme has several drawbacks. For example, if $P(A\,|\,b,c) = P(A\,|\,E)$ then the scheme may give unjustified counts yielding a false accuracy. If, for example, $E = (b,c)$, then nothing can be learned on the distribution of $A$, but nevertheless the sample size will be increased by one. See further discussion of this issue in Spiegelhalter and Lauritzen (1990*a*) as well as in Spiegelhalter and Cowell (1992).

A mathematically correct updating of the distributions under our interpretation results in a mixture of Dirichlet-distributions rather than in a single one (a mixture is a linear combination with non-negative coefficients summing to 1). This complicates the calculations intractably - in particular when adapting from the next case where mixtures of Dirichlet-distributions are to be updated. Eventually the process will yet again result in a combinatorial explosion. Instead, the correct distribution is approximated by a single Dirichlet-distribution (keeping the approach of modifying counts). First of all, we want the approximated distribution to have the correct means, and the new set of probabilities $(m_1^*,\ldots,m_k^*)$ is set to be the means of the correct distribution. Secondly, it would be preferable also to give the distribution the correct variances. However, this is not possible since only one free parameter is left, namely the equivalent sample size. Instead, the equivalent sample size is given a value such that the 'average variance'

$$v = \sum_{i=1}^{k} m_i v_i \tag{4}$$

is correct. The resulting scheme, which is used in aHUGIN, is the following: First the means are changed as if a full count was obtained:

$$m_i^* = \frac{m_i s + P(a_i,b,c\,|\,E) + m_i\{1-P(b,c\,|\,E)\}}{s+1} \tag{5}$$

The last term may be understood so that it distributes the probability that $B$ and $C$ are not in states $(b,c)$ over the $a_i$'s according to their present probabilities.

Next, the sample size is determined:

$$s^* = \frac{\sum_{i=1}^{k} m_i^{*2}(1-m_i^*)}{\sum_{i=1}^{k} m_i^* v_i^*} - 1 \tag{6}$$

where $v_i^*$ is the variance of $p_i$ in the mixture (the formulae may be found in Spiegelhalter and Lauritzen (1990*a*)). The new counts are $s^* m_i^*$.

## 3 Features of aHUGIN

The program aHUGIN, which is currently under implementation, is an extension of HUGIN (Andersen *et al.* 1989). HUGIN is a shell which allows the user to edit CPNs over finite state variables, and when the CPN is specified, HUGIN creates a runtime system for entering findings and updating probabilities of the variables in the network.

In aHUGIN each variable may be declared to be in adaptation mode. If, for example, the variable $A$ with states $a_1,\ldots,a_n$ has parents $B,\ldots,C$, then the conditional probability table $P(A|B,\ldots,C)$ is modified by declaring $A$ of adaptation type. The table is interpreted as a contingency table such that for each parent configuration $b,\ldots,c$, the set $P(A|b,\ldots,c)$ is interpreted as a set of frequencies based on a sample of cases. Therefore the user will for each parent configuration be prompted to specify EQUIVALENT SAMPLE SIZE. The larger the ESS, the more conservative the adaptation will be. The default value of ESS is $5k$, where $k$ is the number of states in $A$.

Alternatively the user will be asked to specify an interval for each of the probabilities in the conditional probability tables. These intervals will then be translated to sample sizes using the equivalent of (3). The ESS used for the given parent configuration will now be chosen as the minimum of the translated sample sizes for the individual entries.

### 3.1 Fading

Variables in adaptation mode have an extra feature, ***fading***, which makes them tend to ignore things they have learnt a long time ago, considering them as less relevant. Each time a new case is taken into account, the equivalent sample size is discounted by a fading factor $q$, which is a real number less than one but typically close to one. From the expression (1) for the Dirichlet density, it is seen that the fading scheme essentially corresponds to flattening the density by raising it to the power $q$, known as power-steady dynamic modelling (Smith 1979; Smith 1981).

If $s$ is the initial ESS, then the maximal ESS after adaptation from a case is $qs + 1$. Running $n$ cases will result in a maximal ESS of

$$q^n s + \frac{1 - q^n}{1 - q}$$

This gives that $1/(1 - q)$ is the maximal sample size in the long run.

Therefore the user is given the choice between ACCUMULATING (fading factor 1) and FADING. If fading is selected, the user is prompted for MAXIMAL SAMPLE SIZE, MSS, and the fading factor is then computed as (MSS - 1)/MSS. Default value is $100m$, where $m$ is the number of entries in the table.

*Note:* The result of fading is not only that the sample size is reduced. Consider namely an entry with count $\alpha$ and with sample size $s$, and suppose that retrieval of a case results in an increase of the sample size by 1 and of the count by $x$. Without fading the ratio between counts from present and past is $x/\alpha$, but with fading the ratio is $x/q\alpha$. This tells us that with fading the present counts are given more weight. This can also be seen by assuming that the entry will never receive more counts. Without fading the probability will vanish at the speed of $\alpha/(s+n)$ while with fading, the speed of vanishing is in the order of $\alpha/(s + q^{-n})$.

### 3.2 Runtime mode

The adaptation starts with the CPN in the initial configuration. Findings are entered, and when all information on the case has been entered, the adaptation takes place changing the tables for the variables of adaptation type.

At any time between two cases the user can choose to change the adaptation type of any variable. When the adaptation type of a variable has been changed, the user is prompted for possible missing information on EQUIVALENT SAMPLE SIZE and MAXIMAL SAMPLE SIZE. In the case of a change from accumulating to fading the EQUIVALENT SAMPLE SIZE is kept but the MAXIMAL SAMPLE SIZE provided by the user will gradually claim its influence.

## 4 Experiments with aHUGIN

To investigate the strengths and limitations of aHUGIN, a series of experiments were carried out. The investigation was designed as a complete factorial simulation experiment on the now classical "Chest clinic" example (Figure 2) originating from Lauritzen and Spiegelhalter (1988). Each experiment simulates

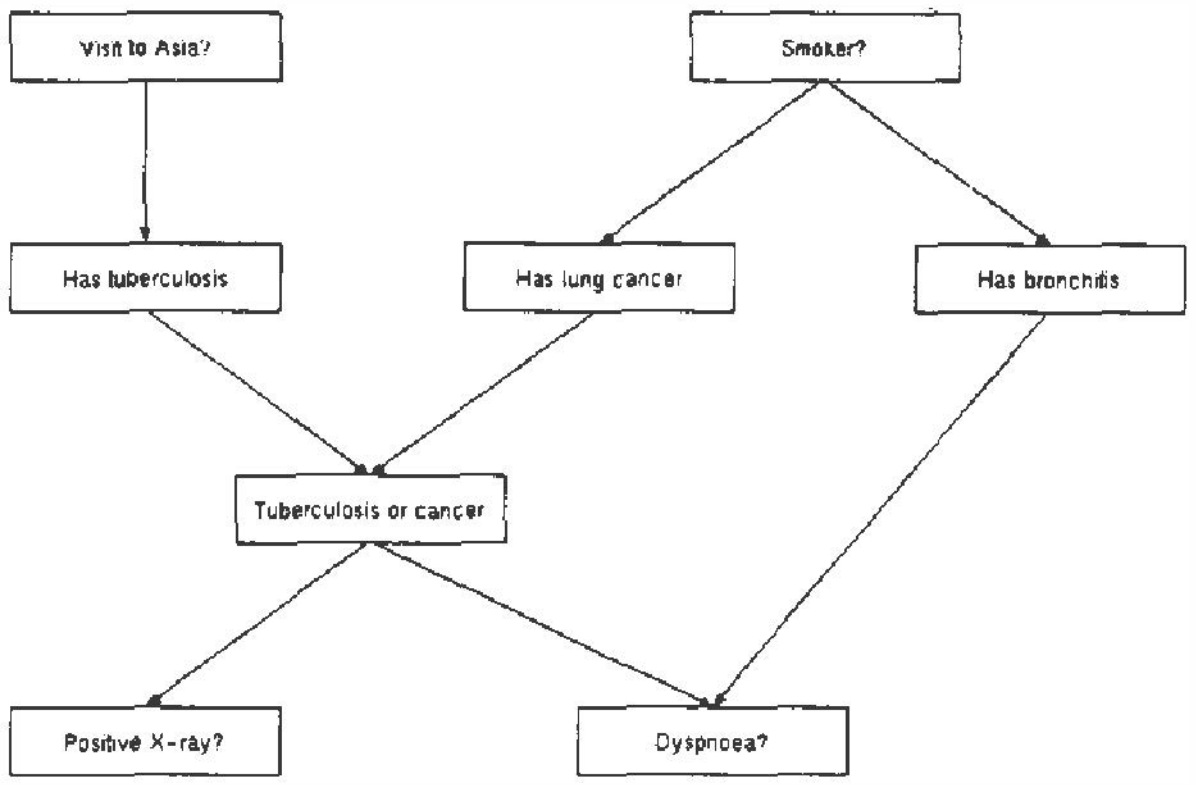

Figure 2: The "Chest clinic" example.

10,000 cases, and four factors, denoted R, O, P and L, are considered. Three random samples (R) are generated from

R1: Probabilities close to the original ones.

R2: Probabilities very different from the original ones.

R3: Probabilities "drifting over time", starting as the original ones.

To control differences due to chance variations, the samples are reused. Thus, for example, all experiments with probabilities as in R1 are based on identical data.

Two different observational schemes (O) are investigated, the first one is mainly included for control purposes.

O1: Complete observations.

O2: Data observed only on the variables "Visit to Asia?", "Smoker?", "Positive X-ray?" and "Dyspnoea?".

The P factor describes different weights on the prior distributions, expressed as varying equivalent sample sizes. Two cases are considered

P1: Low precision, ESS = 10.

P2: High precision, ESS = 100.

Finally, three different learning schemes (L) are investigated

L1: All variables except "Tuberculosis or cancer" in accumulating mode.

L2: As L1 for the first 1000 cases, then the mode is changed to fading, with long memory (MSS = 1000).

L3: As L2, but with short memory (MSS = 100).

"Tuberculosis or cancer" is always in fixed mode as it is a pure logical transition. As can be seen, the whole investigation consists of $3 \times 2 \times 2 \times 3 = 36$ experiments. For each experiment a plot is generated, showing the current value of the conditional probabilities after each case has been processed, together with approximate 95% posterior probability intervals.

### 4.1 Results in accumulating mode

These experiments are very similar to those performed by Spiegelhalter and Cowell (1992). However, we allow uncertainty on all conditional probability tables. In general our results show the same pattern. For complete data the correct values are obtained quite fast, and the influence of the initial specifications vanishes after a few hundred cases.

Figures 3 (a) and 4 (a) show an interesting phenomenon when learning from incomplete data (O2). In these experiments, it can only be observed from the given data that a majority of smokers suffer from dyspnoea (shortness of breath). It can not be inferred from the data whether this correlation is due to the presence or absence of bronchitis. In the first experiment, where all variables are in accumulating mode, the frequency of bronchitis is overestimated (Figure 3 (a)). To compensate for this, the conditional probability for dyspnoea given bronchitis and none of the other diseases, is underestimated (Figure 4(a)). Thus the correlation between what can actually be observed in the data is determined correctly, but the intermediate explanation is slightly incorrect.

From these experiments we conclude, not surprisingly, that the method has difficulties learning about concepts on which data are indirect. In such situations the system relies strongly on prior knowledge. This conclusion was also reached by Spiegelhalter and Cowell (1992).

### 4.2 Results in fading mode

Figures 3 (b)-(c) and 4 (b)-(c) display the results for the same experiments as in Figures 3 (a) and 4 (a), but with the variables changed to fading with long memory after the first 1000 cases. The same effect on estimating intermediate variables can be observed. Note also, that the two curves vary synchronously. Most probably this is a result of variations in frequencies due to the random sample.

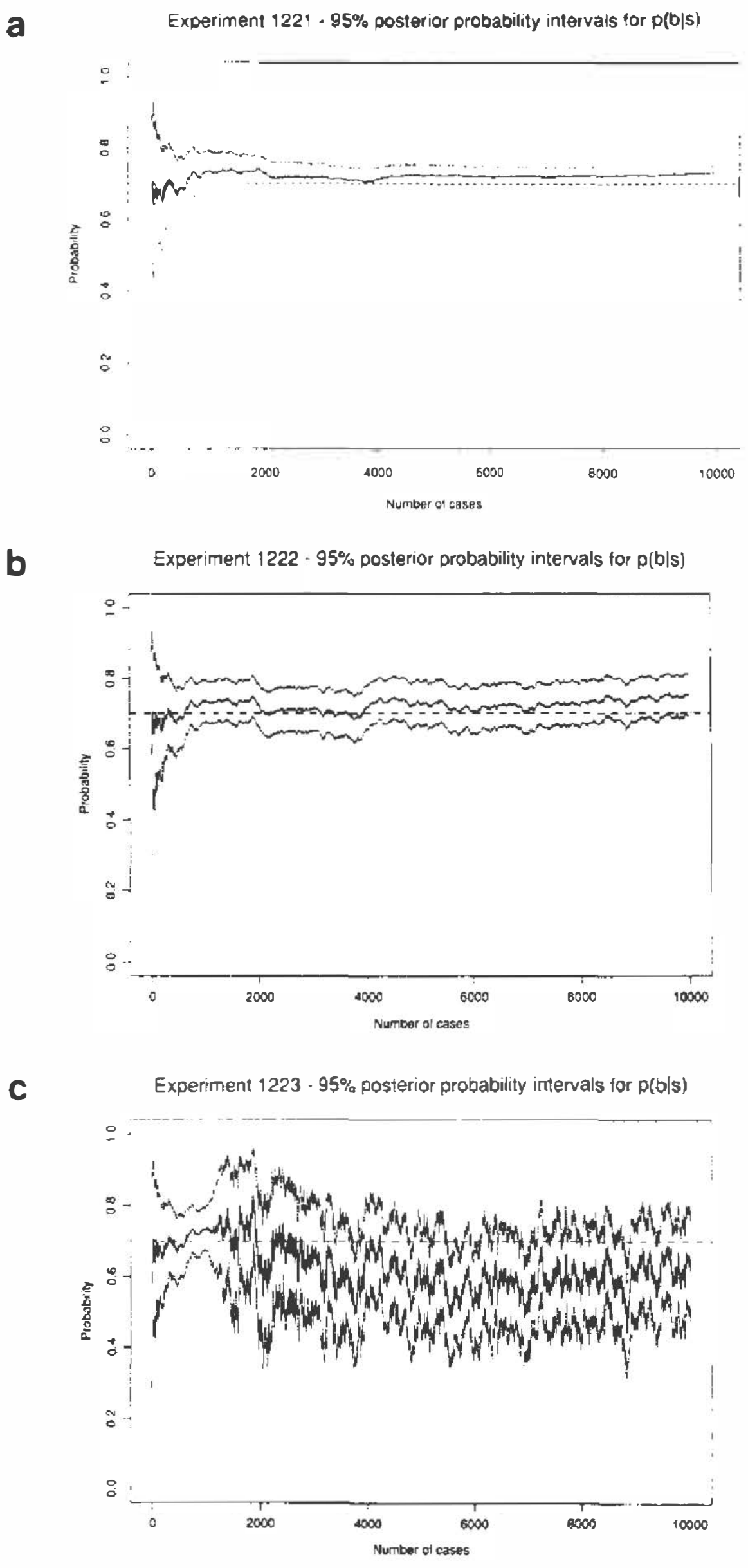

Figure 3: Experiments with incomplete data. The conditional probability of bronchitis given the patient is a smoker is learnt in (a) accumulating mode; (b) fading mode with long memory; (c) fading mode with short memory.

In the third experiment (Figures 3 (c) and 4 (c)) the maximal sample sizes are reduced to 100. This experiment reveals the limit of the applicability of the fading mode. Figure 4 (c) shows that the data are best explained by assuming that all patients with bronchitis suffer from dyspnoea. To maintain the consistency with the data, the frequency of smokers suffering from bronchitis is underestimated accordingly. This pattern is general for fading with short memory for high and low probabilities. We conclude that special attention must be directed towards systematically missing data and the choice of MSS if such variables are fading. Figure 5 shows a series of experiments with a declining probability of being a smoker. The first 1000 cases are identical for the three plots, the variable being in accumulating mode. In Figure 5 (a) the variable remains in this mode and it is seen how the probability is becoming increasingly conservative as the ESS increases.

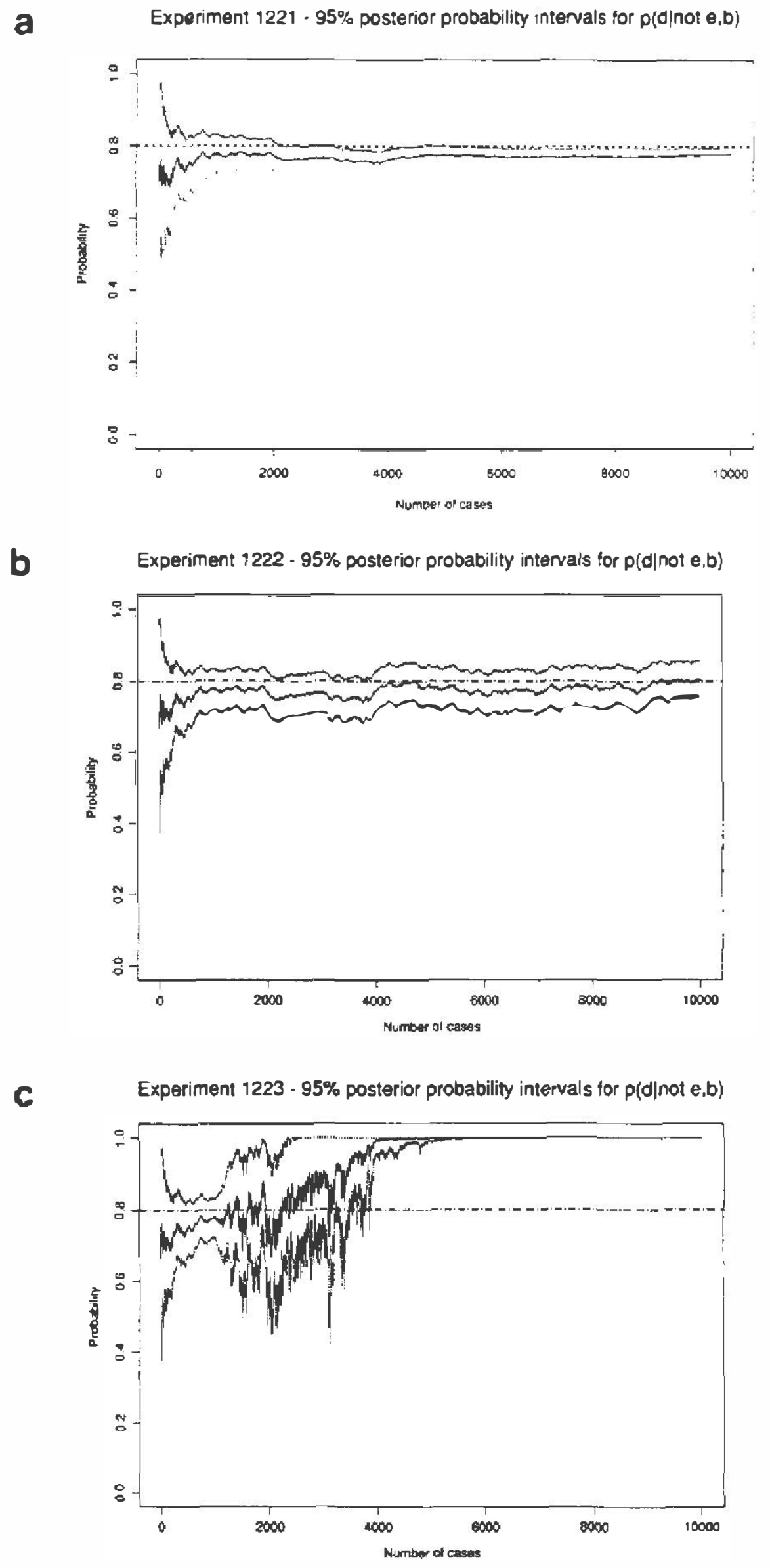

Figure 4: The same experiment as in Figure 3 but for "Dyspnoea" given the patient has bronchitis but none of the other diseases.

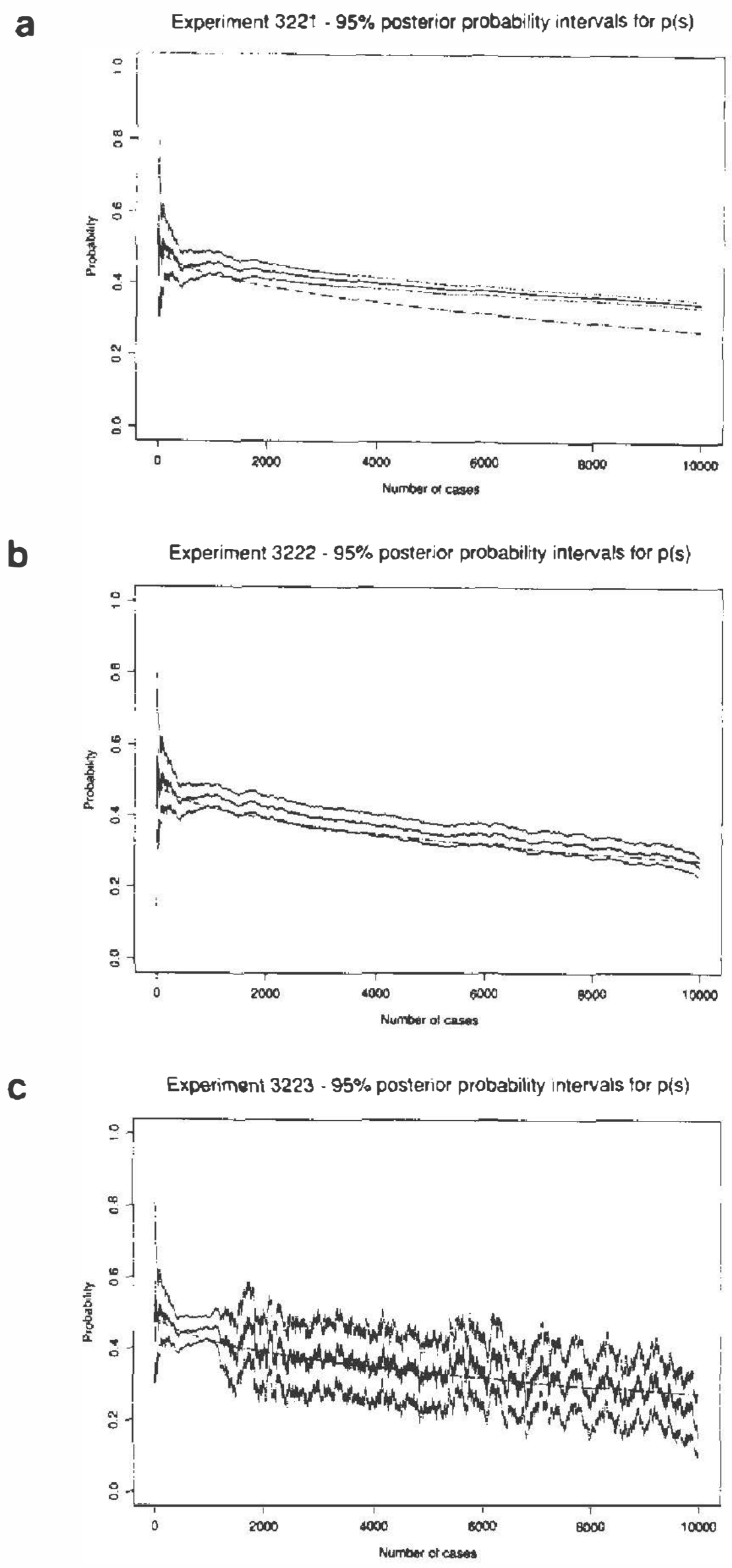

Figure 5: Learning about a declining probability of being a smoker.

In Figure 5 (b) the variable is changed to fading with long memory (MSS = 1000) after the first 1000 cases. This increases the dynamic behaviour of the system and an almost correct adaptation is obtained. Decreasing the MSS to 100 (Figure 5 (c)) increases the dynamic behaviour further, resulting in stronger fluctuations around the correct value. The general experience is that the MSS should not be set too low, and that the experiments confirm the expected behaviour of aHUGIN.

To summarize, aHUGIN seems to be able to adapt to changing environments, thereby extending HUGIN with a valuable functionality. However, special attention must be directed to the choice of MSS and to variables with systematically missing data.